\documentclass[10pt, a4paper]{article}
\usepackage[]{lrec-coling2024}
\usepackage[extra]{tipa}

\title{nEMO: Dataset of Emotional Speech in Polish}

\name{Iwona Christop} 

\address{Adam Mickiewicz University\\
        Faculty of Mathematics and Computer Science\\
        Uniwersytetu Pozna{\'n}skiego 4, 61-614 Pozna{\'n}, Poland \\
        christop@wmi.amu.edu.pl}

\abstract{
Speech emotion recognition has become increasingly important in recent years due to its potential applications in healthcare, customer service, and personalization of dialogue systems. However, a major issue in this field is the lack of datasets that adequately represent basic emotional states across various language families. As datasets covering Slavic languages are rare, there is a need to address this research gap. This paper presents the development of nEMO, a novel corpus of emotional speech in Polish. The dataset comprises over 3 hours of samples recorded with the participation of nine actors portraying six emotional states: anger, fear, happiness, sadness, surprise, and a neutral state. The text material used was carefully selected to represent the phonetics of the Polish language adequately. The corpus is freely available under the terms of a Creative Commons license (CC BY-NC-SA 4.0).
 \\ \newline  }

\begin{document}

\maketitleabstract

\section{Introduction}

Automatic speech recognition is used in various aspects of life, ranging from customer service systems to virtual assistants and chatbots. As speech is the most natural form of communication for humans, human-computer interaction systems strive to minimize written interfaces.

Incorporating speech emotion recognition in dialogue systems may not seem like an obvious next step. However, this development could significantly refine the personalization of virtual assistants, enabling them to respond appropriately to the emotional state of the user. The impersonal nature of current dialogue systems often results in user discouragement \citep{serTwoDecades2018}.

The potential applications of emotion recognition extend beyond customer service. For example, in the context of emergency calls, the ability to distinguish emotions such as fear or sadness could be crucial in assessing the authenticity of the call. Similarly, \citet{Abbaschian2021DeepLT} demonstrated its utility within therapy sessions, where the analysis of a patient's emotions could offer insights into their psychological state.

Although speech emotion recognition is acknowledged as a critical area of research, the field still faces challenges, most notably the scarcity of high-quality, diverse datasets. Most available corpora contain simulated emotional expressions that do not accurately reflect human emotions \citep{Abbaschian2021DeepLT}. Additionally, the development of resources for various natural language processing and speech processing tasks often overlooks low-resource languages.

Polish being the second most spoken language within the Slavic language family exemplifies a significant gap in resources for speech emotion recognition. This study aims to bridge this gap with the introduction and analysis of the nEMO dataset, a novel corpus of Polish emotional speech.

The following section provides an overview of existing emotional speech corpora. Section \ref{sec:methodology} describes the methodology used to create the nEMO dataset, including the selection of linguistic content, emotional states, and participant actors. Section \ref{sec:final-corpus} provides an overview of the final dataset. Section \ref{sec:evaluation} describes the evaluation of the dataset using various machine learning techniques. Finally, section \ref{sec:conclusions} provides conclusions and considerations for future work.

\section{Related Work}\label{sec:related-work}

\begin{table*}[!htb]
\begin{center}
\begin{tabular}{|l|l|l|l|l|p{0.25\textwidth}|} \hline 
    & & & \textbf{Number of} & \textbf{Number of} & \\
    \textbf{Dataset} & \textbf{Language} & \textbf{Type} & \textbf{samples} & \textbf{speakers} & \textbf{Emotions} \\ \hline 
    \textbf{CREMA-D} & English & simulated & 7,742 & 91 & anger, disgust, fear, happiness, neutral, sadness \\ \hline
    \textbf{EMO-DB} & German & simulated & 700 & 10 & anger, boredom, disgust, fear, happiness, neutral, sadness \\ \hline
    \textbf{IEMOCAP} & English & semi-natural & 1,150 & 10 & anger, disgust, excitement, fear, frustration, happiness, neutral, sadness, surprise \\ \hline
    \textbf{RAVDESS} & English & simulated & 2,496 & 24 & anger, calm, disgust, fear, happiness, neutral, sadness, surprise \\ \hline
    \textbf{TESS} & English & simulated & 2,800 & 2 & anger, disgust, fear, happiness, neutral, sadness, surprise \\ \hline
    \textbf{URDU} & Urdu & natural & 400 & 38 & anger, happiness, neutral, sadness \\ \hline
\end{tabular}
\caption{An overview of popular emotional speech datasets, highlighting language, type, number of samples and speakers, and emotional states covered.}
\label{tab:related-work}
\end{center}
\end{table*}

The reliability of speech emotion recognition systems depends on the quality of data used for training. Although finding a corpus for Slavic languages poses a challenge, there are valuable datasets available for other language families. This indicates a global interest and effort in the development of emotional speech datasets.

The datasets of emotional speech can generally be classified into three categories based on the method used to acquire recordings: natural, semi-natural, and simulated.

Natural datasets contain authentic emotional states instead of exaggerated expressions. These datasets are typically sourced from YouTube videos, television series, and call center recordings. While this approach prevents overfitting to artificial emotions, it comes with other complications. The continuity and dynamics of the recordings can complicate emotion detection, and multiple emotional states may be present simultaneously. Additionally, we have no control over the selection of presented emotions. Furthermore, the subjective nature of emotions requires manual annotation, which can vary between individuals. Finally, such content may be restricted by copyrights \citep{Abbaschian2021DeepLT}.

In the case of semi-natural datasets, actors attempt to portray emotions based on a given script, relying on their own perception of the content. While this approach strives to align with natural speech, the expressions may seem unnatural due to the specified context. Similar to natural datasets, the interpretation and portrayal of emotional states are actor-dependent, and manual labeling is required \citep{Abbaschian2021DeepLT}.

For simulated datasets, professional actors read the same text while portraying different emotions. This approach allows for the definition of an arbitrary number of discrete emotional states, simplifying comparative analysis across different corpora and ensuring full control over copyrights. However, a major limitation of such datasets is that the enacted emotions may seem unnatural and distorted from conversational speech \citep{Abbaschian2021DeepLT, serTwoDecades2018}.

Although natural datasets may seem like the best choice due to their authentic representation of human emotions, the majority of existing resources utilize the simulated approach because of its vast advantages. The selection of the approach is usually followed by determining the spectrum of emotions to be included. Existing datasets consist of portrayals of various emotional states, including anger, fear, happiness, sadness, and neutral state. The curation of linguistic material that accurately reflects the phonetics of the targeted language is the final essential step.

Table \ref{tab:related-work} presents a comparative overview of several well-known emotional speech datasets: CREMA-D \citep{crema-d}, EMO-DB \citep{emo-db}, IEMOCAP \citep{iemocap}, RAVDESS \citep{ravdess}, TESS \citep{tess}, and URDU \cite{urdu}. It summarizes each dataset by providing information on its language, category, total number of samples, number of contributing speakers, and the range of emotions covered. This comparison provides an understanding of the diversity of existing resources in the field of speech emotion recognition.

\section{Methodology}\label{sec:methodology}

\begin{table*}[!htb]
\begin{center}
\begin{tabular}{|c|c|c|p{0.23\textwidth}|p{0.23\textwidth}|} \hline 
    \textbf{Phoneme} & \textbf{Word} & \textbf{Word in IPA} & \textbf{Sentence in Polish} & \textbf{Sentence in English} \\ \hline
    \textipa{\~i} & zima & [\textipa{\textprimstress\textctz\~ima}] & Zima to czas je\.{z}d\.{z}enia na sankach. & Winter is the time for sledding. \\ \hline
    \textipa{b} & baza & [\textipa{baza}] & Baza wojskowa jest strategicznym punktem dla armii. & The military base is a strategic point for the army. \\ \hline
    \textipa{\textroundcap{\texttslig}\textsuperscript{j}} & racja & [\textipa{\textprimstress ra\textroundcap{\texttslig}\textsuperscript{j}ja}] & Racja, ten pomys\l{} jest \mbox{najlepszy}. & Right, this idea is the best. \\ \hline
    \textipa{s\textsuperscript{j}} & pasja & [\textipa{pas\textsuperscript{j}ja}] & Pi\l{}ka no\.{z}na to moja pasja. & Soccer is my passion. \\ \hline
    \textipa{\r*l\textsuperscript{j}} & rzemie\'{s}lniczy & [\textipa{\textsecstress\textyogh\~\textepsilon m\textsuperscript{j}j\.\textepsilon\textctc\r*l\textsuperscript{j}\textprimstress\textltailn i\textroundcap{\textteshlig}\textbari}] & Na targach rzemie\'{s}lniczych mo\.{z}na zobaczy\'{c} wiele unikatowych produkt\'{o}w. & At craft fairs, you can see many unique products. \\ \hline
\end{tabular}
\caption{Representative sentences from the nEMO datasets that demonstrate the phonetic diversity of Polish, along with their English translations.}
\label{tab:sentences}
\end{center}
\end{table*}

\subsection{Emotional States}

We chose the simulated approach for the development of the nEMO dataset due to its numerous advantages. This method allowed us to define emotions in a distinct and categorical manner, while also providing control over copyrights.

The nEMO dataset focuses on six basic emotions: anger, fear, happiness, sadness, surprise, and neutral state. These emotions were selected due to their differences in valence, arousal, and dominance, which facilitates their discrimination \citep{plutchikKellerman}. Furthermore, these emotional states are frequently depicted in existing corpora, allowing for an increase in the number of samples for basic emotions.

\subsection{Linguistic Content}

In the development of the nEMO dataset, a simulated approach was used where each actor was required to record the same set of utterances for each of the six different emotional states. To ensure optimal duration of the recording sessions, an effort was made to minimize the number of utterances as much as possible. Additionally, the linguistic content was prepared to sufficiently represent the phonetics of the Polish language.

The initial step in preparing linguistic material involved identifying 90 uncommon phonemes present in the Polish language. For each phoneme, a word in which it appears was selected and used in a sentence. This resulted in the creation of 90 sentences, each containing at least one uncommon phoneme.

It should be noted that all phrases are semantically correct and could be used in everyday conversations. Furthermore, the emotional neutrality of the sentences was not emphasized.

Table \ref{tab:sentences} shows examples of the sentences used in the recordings.

\subsection{Actors}

Nine actors, four female and five male, were involved in the development of the nEMO dataset. They ranged in age from 20 to 30 and were all native speakers of Polish.

Three of the participants were qualified voice actors. To avoid exaggerated emotional expressions, which are common among theatrical performers, individuals without formal acting training were also included.

The dataset was balanced by strategically incorporating non-professional speakers to reduce the potential for exaggerated emotional portrayals. Although concerns were raised about the adequacy of emotional expression by non-actors, analysis revealed no significant differences between professionals and amateurs, both audibly and in spectral data analysis. To maintain the dataset's integrity, we excluded any recordings that raised doubts about their emotional authenticity through human evaluation.

During the recording sessions, actors were given explicit instructions to focus on depicting a single emotional state at a time. Feedback was provided constantly to all participants, particularly aiding those without professional experience. This guidance often included encouragement to use non-verbal cues, such as facial expressions and gestures, to enhance the authenticity of emotional expression. For example, actors were instructed to smile when portraying happiness, in order to enhance the emotional quality of the recordings.

All participants provided consent for the use and distribution of their voices in the form of audio recordings. This agreement covers the rights to use, preserve, process, and reproduce the audio content. It also includes provisions for unrestricted distribution and public availability under the terms of a Creative Commons license (CC BY-NC-SA 4.0), ensuring ethical and legal use of the dataset in subsequent research.

\begin{table*}[!htb]
\begin{center}
\begin{tabular}{|p{0.2\textwidth}|cccccc|c|} \hline 
    \textbf{Emotion} & \textbf{Anger} & \textbf{Fear} & \textbf{Happiness} & \textbf{Neutral} & \textbf{Sadness} & \textbf{Surprise} & \textbf{Total} \\ \hline
    \textbf{Number of samples} & 749 & 736 & 749 & 809 & 769 & 669 & 4,481 \\ 
    \textbf{Average length [s]} & 2.26 & 2.49 & 2.34 & 2.56 & 2.69 & 2.46 & 2.47 \\ 
    \textbf{Total length [h]} & 0.47 & 0.52 & 0.48 & 0.57 & 0.57 & 0.45 & 3.07 \\ \hline
\end{tabular}
\caption{Quantitative analysis of the nEMO dataset per emotion: number of samples, average sample duration, and total length.}
\label{tab:final-dataset}
\end{center}
\end{table*}

\subsection{Recording Setup}

The recordings were conducted in a home setting to better reflect a natural environment. Each recording session involved one actor and lasted approximately two hours, excluding post-processing or sample evaluation.

The utterances were captured using Mozos MKIT-900 Pro, a cardioid condenser microphone with a 192 kHz sampling rate, \textasciitilde42 dB sensitivity, and frequency response from 100 Hz to 18 kHz. The recording equipment also included a sponge and pop filter to eliminate background noise and explosive consonant utterances. As a result, there was no need for any post-processing, as the quality of the recordings was not affected by any external noise. All samples were normalized to a peak amplitude of 0 dB and downsampled to a 24 kHz sampling rate with 16 bits per sample.

\section{Final Dataset}\label{sec:final-corpus}

The nEMO dataset underwent human evaluation, and only recordings that accurately captured the intended emotional state were included. The resulting dataset contains 4,481 audio recordings, a total of more than three hours of speech. Table \ref{tab:final-dataset} shows the distribution of samples per emotional state.

The final version of the nEMO dataset includes a comprehensive set of attributes. These attributes consist of an audio sample, a label corresponding to the emotional state, and both original and normalized transcriptions of the utterance. Additionally, metadata related to the speaker's ID, gender, and age are included.

The nEMO dataset is available as a single entity and is not divided into predefined training and test splits. This approach allows researchers and developers to customize the splits according to their specific needs.

The dataset's diverse attributes make it a valuable asset for a broad spectrum of speech processing tasks. It was primarily designed to facilitate speech emotion recognition and contains recordings annotated with one of six discrete emotional states: anger, fear, happiness, sadness, surprise, or neutral. Additionally, the dataset includes detailed metadata on the speaker, making it suitable for various audio classification tasks. Furthermore, including both orthographic and normalized transcriptions for each sample enhances the dataset's utility for automatic speech recognition tasks. The linguistic content has been carefully selected to showcase a broad spectrum of the phonetics of the Polish language. Additionally, the emotional speech recordings, complemented by transcriptions, lay the groundwork for the development of text-to-speech systems capable of generating emotional speech.

\section{Evaluation}\label{sec:evaluation}

To ensure accurate representation of designated emotional states, three experimental evaluations were conducted using machine learning algorithms. The dataset was randomly divided into training and test sets at an 80:20 ratio. It is important to note that the splits were not speaker-independent.

The experiments utilized three classifiers: Support Vector Machine (\textbf{SVM}), Logistic Regression, and Random Forest. These classifiers leveraged Mel-frequency cepstral coefficients (\textbf{MFCCs}) as input features. From each audio recording, 20 MFCCs were computed and then averaged to obtain a one-dimensional feature vector. This approach, based on basic features and machine learning methods, was chosen to provide interpretable results of the experiments.

\begin{figure}[!ht]
\begin{center}
\includegraphics[width=\columnwidth]{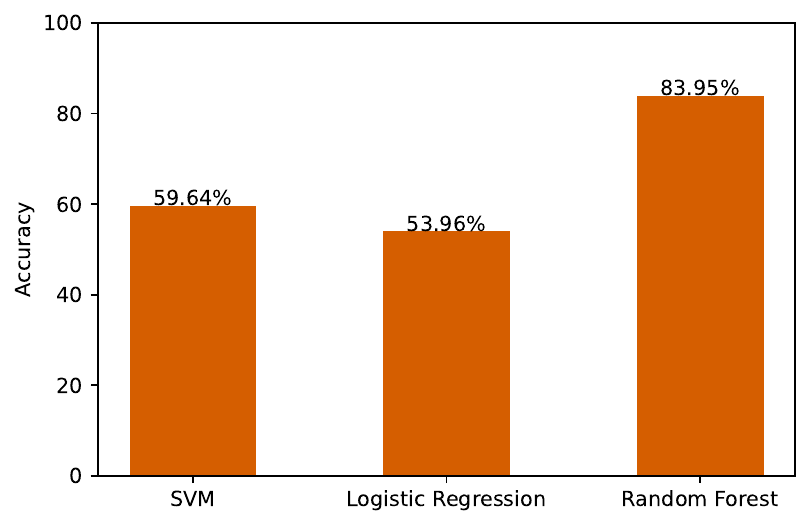} 
\caption{Performance metrics of SVM, Logistic Regression, and Random Forest classifiers on the nEMO dataset.}
\label{fig:accuracy}
\end{center}
\end{figure}

\begin{figure*}[htb!]
    \centering
    \includegraphics[width=\textwidth]{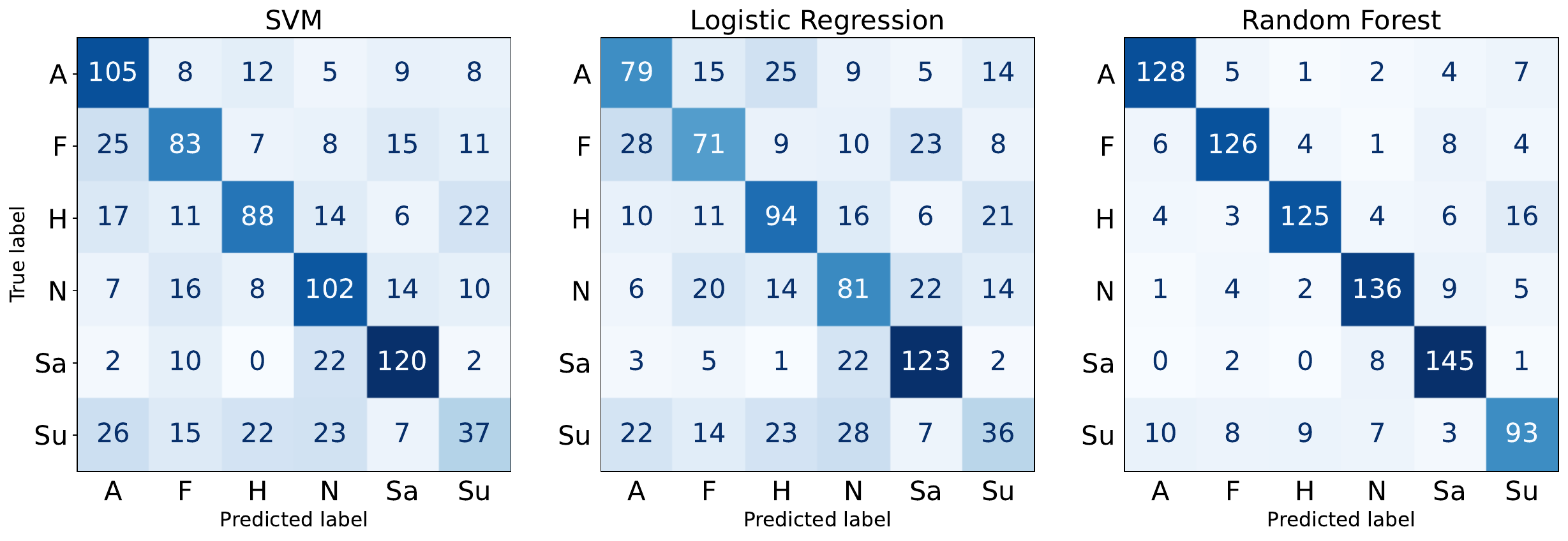} 
    \caption{Comparison of confusion matrices for SVM, Logistic Regression, and Random Forest classifiers on the nEMO dataset.}
    \label{fig:confusion-matrix}
\end{figure*}

The performance of the classifiers is shown in Figure \ref{fig:accuracy}, indicating satisfactory outcomes. Random Forest achieved the highest accuracy at 83.95\%. The confusion matrices for all classifiers are shown in Figure \ref{fig:confusion-matrix}.

The SVM classifier achieved 59.64\% accuracy and a well-balanced precision-recall ratio with a macro F1 score of 58.17\%. However, the classifier exhibited a tendency for errors in recognizing the emotional state of \textit{surprise}. The notably lower recall rate for this emotional state suggests a need for reevaluation of these particular samples.

The Logistic Regression model produced comparable results to the SVM in terms of accuracy and macro F1 score. It also had problems in classifying \textit{surprise} accurately.

The errors made by both models were mainly related to misclassification between emotions with similar levels of arousal, highlighting the complexity of the dataset and the nuanced challenge it poses. In contrast, the Random Forest classifier achieved the highest accuracy and demonstrated consistent performance across all emotional states.

\section{Conclusions}\label{sec:conclusions}

This paper presents the development and evaluation of the nEMO dataset, a novel Polish emotional speech corpus. This dataset, which contains recordings of nine actors delivering 90 sentences in six different emotional states, consists of 4,481 samples, representing over three hours of speech data. It includes a wide range of metadata fields, including raw audio recordings, emotional state labels, speaker gender and age, and utterance transcriptions, thereby extending its utility beyond its primary goal for various speech processing tasks.

The robustness of the dataset was evaluated using three machine learning methods: SVM, Logistic Regression, and Random Forest. The satisfactory performance of these classifiers proves the effectiveness of the dataset in capturing the nuanced phonetics of Polish emotional speech, highlighting its potential as a valuable asset for speech emotion recognition research.

Given the availability of specialized corpora in the field of speech emotion recognition, the nEMO dataset is an important resource aimed at bridging this gap. Future work for the corpus includes extensive development, including further human evaluation and the inclusion of additional recordings from more actors to further increase its diversity and applicability.

The nEMO dataset is available under a Creative Commons license (CC BY-NC-SA 4.0) on platforms such as the Hugging Face\footnote{\url{https://huggingface.co/datasets/amu-cai/nEMO }} website and GitHub\footnote{\url{https://github.com/amu-cai/nEMO}} to enable further research and development in the field of speech emotion recognition. By making nEMO publicly available, we hope to encourage progress in the field and the creation of effective emotion recognition systems.

\section{Acknowledgments}

The author would like to thank all the actors whose voices are the essence of the nEMO dataset. Their commitment and dedication were instrumental in capturing the diverse emotional states that this corpus seeks to represent. Their contribution is the foundation of this research.

Special thanks are due to Dr. Marek Kubis, whose invaluable support and insightful guidance were crucial throughout the research and development of this project.

\nocite{*}
\section{Bibliographical References}\label{sec:reference}
\bibliographystyle{lrec-coling2024-natbib}
\bibliography{bibliography}

\end{document}